# Variance Based Moving K-Means Algorithm

Vibin Vijay
Dept. of Electronics and Communication Engineering
National Institute of Technology Warangal
Warangal, India
vvibin@student.nitw.ac.in

Raghunath V P
Dept. of Electronics and Communication Engineering
National Institute of Technology Warangal
Warangal, India
vraghunath@student.nitw.ac.in

Amarjot Singh
School of Engineering Science
Simon Fraser University
Burnaby, Canada
asa168@sfu.ca

SN Omkar
Dept. of Aerospace Engineering
Indian Institute of Science
Bangalore, India
omkar@aero.iisc.ernet.in

*Abstract*— Clustering is a useful data exploratory method with its wide applicability in multiple fields. However, data clustering greatly relies on initialization of cluster centers that can result in large intra-cluster variance and dead centers, therefore leading to sub-optimal solutions. This paper proposes a novel variance based version of the conventional Moving K-Means (MKM) algorithm called Variance Based Moving K-Means (VMKM) that can partition data into optimal homogeneous clusters, irrespective of cluster initialization. The algorithm utilizes a novel distance metric and a unique data element selection criteria to transfer the selected elements between clusters to achieve low intra-cluster variance and subsequently avoid dead centers. Quantitative and qualitative comparison with various clustering techniques is performed on four datasets selected from image processing, bioinformatics, remote sensing and the stock market respectively. An extensive analysis highlights the superior performance of the proposed method over other techniques.

## I. INTRODUCTION

Clustering aims at grouping unlabeled data elements with high similarity into clusters based on any measure obtained solely from the data. These methods have been widely used in different investigative areas such as face detection [10], bioinformatics [9, 1, 14], market analysis [2] etc. Clustering has been extensively used to detect faces using skin extraction [5] while bioinformatics researchers utilized cluster analysis to build gene groups with related patterns and develop homologous sequences of genes [14]. Furthermore, market researchers took advantage of clustering techniques to segment multivariate survey data to better understand the relationships between different groups of consumers [2].

A rich collection of methods has been proposed in the past to group congruent data elements. K-Means (KM) clustering is the most popular method that divides the data into disjoint clusters using the Euclidean distance between a data element and the center of the aligned cluster. Despite the algorithm's wide use and strong advantages, the algorithm: (i) is sensitive to the cluster center initialization that can result in clusters with large intra-cluster variance and dead centers, further leading in bad data division (ii) has low capability to overcome locally optimum solutions.

A number of algorithms have been proposed to overcome the above-mentioned disadvantages including but not limited to Moving K-means (MKM), Adaptive Moving K-Means (AMKM), Fuzzy C-Means (FCM), Kernel K-Means and Enhanced Moving K-Means (EMKM-1, EMKM-2). Moving K-Means (MKM) algorithm [8] addresses the issue of trapped or dead (undesired) centers by constantly checking each cluster's fitness (homogeneity measure) with the objective of maintaining a comparable fitness value for all the clusters. The algorithm avoids clusters with dead centers, however, may group dissimilar elements within a cluster resulting in high intra-cluster variance that can result in bad cluster division. Another version of MKM known as Adaptive Moving K-Means (AMKM) divides data elements into clusters by maintaining minimum intra-cluster variance. This method comparatively reduces the intra-cluster variance but is less effective in avoiding the undesired centers at local minima leading to poor data division. Fuzzy C-Means (FCM) algorithm proposed by Dunn et al. [3] iteratively makes a soft assignment of each data element with a certain membership to all the clusters. This somewhat reduces the sensitivity of the algorithm to initialization allowing it to partition data into clusters with low variance. However, the membership associated with each data element can result in undesired locally optimum solutions. Kernel K-Means clustering [6] has been extensively used to identify clusters that are non-linearly separable in the input space. The use of kernels allows mapping implicitly non-linear data into a high dimensional linear data space, resulting in homogenous clusters with low intra-cluster variance. However, cluster in the kernel space still depend upon initialization conditions and a bad cluster center initialization can result in dead centers. In addition, Kernel K-Means can be computationally expensive for complicated kernels and result in a divergent process.

Enhanced Moving K-Means (EMKM-1, EMKM-2) [12] resolves the above-mentioned problems and produces homogeneous clusters by moving data elements outside half the circle radius, concentered at the cluster center with the highest fitness, to the nearest clusters. Data elements are further transferred from the nearest to the smallest clusters to avoid dead centers and to produce clusters with comparable fitness. This process can produce clusters with high intra-cluster variance as the elements transferred from outside the half radius boundary may have a high correlation with the parent clusters, especially if the parent cluster is dense and compact (small intra-cluster variance).

In this paper, we propose an improved version of the MKM algorithm, with enhanced data transferring steps, known as Variance Based Moving K-means (VMKM). The proposed algorithm can generate homogeneous clusters with low intra-cluster variance without dead centers. The contributions of the proposed algorithm are mentioned below:

- Statistical distance: The proposed algorithm uses Mahalanobis distance to compute correlations of the pixel with its parent and neighbouring clusters. This metric takes into account the correlations of each element with the cluster distribution providing us with a statistically stable measure.
- Variance based selection criteria: The proposed element selection criterion selects the elements with low correlation with their parent cluster and then transfers them to the appropriate neighbouring clusters. This selection criterion results into clusters with low intra-cluster variance and homogeneity as only the elements with the lowest correlation are transferred from the parent cluster, unlike in EMKM where elements outside half radius boundary for each cluster are forcefully transferred.

The rest of the paper is organized as follows. Section II details the proposed VMKM algorithm while Section III compares the performance of the proposed method with seven state of the art clustering techniques on four datasets using mean squared error (MSE). Finally, Section IV draws useful conclusions and presents a direction for future research.

## II. PROPOSED ALGORITHM

The proposed VMKM algorithm employs hard membership function to partition data elements into disjoint clusters by performing four primary steps: (i) First, the cluster centers are randomly initialized by the user (ii) Secondly, Mahalanobis distance metric is used to compute correlation of each element to cluster distributions and assign elements to the cluster with minimum Mahalanobis distance (iii) Thirdly, data elements to be transferred are selected for each cluster using the proposed variance based selection criteria (iv) Finally, data elements are transferred to nearby clusters such that the final clusters have low intra-cluster variance, homogeneity, and no dead centers. Ideal grouping is accomplished if each cluster has a significant number of members, a small difference in fitness and low intra-cluster variance among all the clusters. The algorithm is a generic framework that can be applied to any application. However, we demonstrate it on a few sample images selected from the Berkeley image dataset [4]. Images are chosen to demonstrate the algorithm as they enable us to represent the clusters visually.

The algorithm is applied on a grey scale image $I$ composed of $N$ pixels, required to be grouped into $K$ clusters. The cluster centers are first randomly initialized by the user. The $i^{th}$ pixel, with grey scale intensity value $v_i$ is assigned to the $j^{th}$ cluster $c_j$, where $i = 1, 2, 3...N$ and $j = 1, 2, 3...K$, if and only if it has the minimum Mahalanobis distance compared to the other cluster centers. The Mahalanobis distance d is given by:

$$d(v_i) = (v_i - c_j) \times S^{-1} \times (v_i - c_j)' \qquad (1)$$

where $S$ is the covariance matrix and $S^{-1}$ is the inverse of the covariance matrix. Once the pixels are assigned to each cluster, the cluster center for the $j^{th}$ cluster $c_j$ is updated using:

$$c_j = (1/N) \sum_{i \in c_j} v_i \qquad (2)$$

where $n_j$ is the number of points in the $j^{th}$ cluster. Equation 2 can be similarly used to update remaining clusters. Then, the fitness is calculated for each cluster using the condition in the MKM algorithm, given below:

$$f(c_j) = \sum_{i \in c_j} \|v_i - c_j\|^2 \qquad (3)$$

The cluster with the minimum value of fitness is denoted by $c_s$ and that with the largest value of fitness is denoted by $c_l$. For optimal clustering, the clusters must satisfy the condition derived in MKM algorithm given as:

$$f(c_s) > \alpha_a f(c_l) \qquad (4)$$

Here $\alpha_a$ is a constant value equal to $\alpha_0$ and $\alpha_0$ is a constant with a typical value between $0 < \alpha_0 < 1/3$, found experimentally.

If this condition is not satisfied, data elements with low correlation to the parent cluster that violate Equation 5 are transferred from the largest to the nearest cluster.

$$0 < \sigma_{vl} < 2\sigma_{cl} \qquad (5)$$

Here, $\sigma_{cl}$ is the variance of the largest cluster and $\sigma_{vl}$ is the variance of the pixel belonging to the largest cluster. This step reduces intra-cluster variance. In the next step, the data elements are moved to the cluster with the smallest fitness from its neighbouring cluster in an attempt to avoid *dead centers*. All the points in the neighbouring cluster which violate the condition given in the following Equation 6 are moved to the smallest cluster.

$$0 < \sigma_{vn} < \sigma_{cn} \qquad (6)$$

Here $\sigma_{cn}$ is variance of the nearest neighbouring cluster and $\sigma_{vn}$ is the variance of the pixels belonging to the nearest cluster.

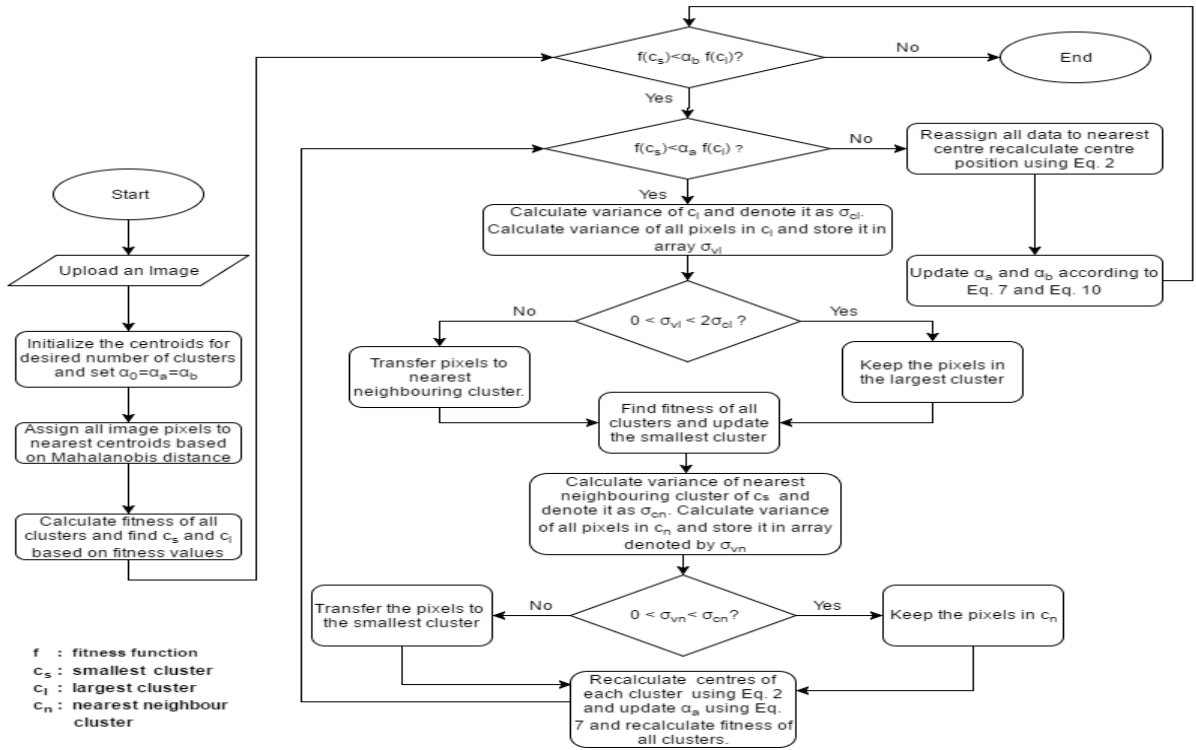

Fig. 1. Illustration of the proposed algorithm using a flowchart that explains each step in detail.

After this step, the fitness of each cluster is re-calculated using Equation 2. The smallest fitness cluster is denoted by $c_s$ and while the largest is denoted by $c_l$ and the value of $\alpha_a$ is updated according to:

$$\alpha_a = \alpha_a - \alpha_a/K. \qquad (7)$$

The process is repeated until Equation 4 is fulfilled. In order to ensure optimal clustering, the process is repeated till the following condition is met by the clusters.

$$f(c_s) > \alpha_b f(c_l) \qquad (8)$$

If this condition is not fulfilled then the whole process is repeated. For each iteration, the constants $\alpha_a$ and $\alpha_b$ are updated according to:

$$\alpha_a = \alpha_0 \qquad (9)$$
$$\alpha_a = \alpha_a - \alpha_a/K. \qquad (10)$$

In Equations 9 and 10, $\alpha_0$, $\alpha_a$ and $\alpha_b$ are initialized as $\alpha_a = \alpha_b = \alpha_0$. Despite the conditions stated in Equations 4 and 8, the algorithm can diverge if any one of them is not satisfied within pre-defined number of iterations.

In order to ensure the convergence, the algorithm is forced to terminate automatically after 100 iterations. The algorithm may terminate before the maximum number of iterations if the conditions in Equations 4 and 8 are met or if after three consecutive iterations, the number of data elements transferred between clusters for each iteration is less than 1% of the total elements. The flow of process is summarized in the flowchart depicted in Fig. 1.

The algorithm described above is demonstrated on an image selected from the Berkley image dataset as shown in Fig. 2. The aim of the convergence is to segment the image in four clusters while iteratively satisfying the conditions detailed in 4 (Nested Loop) to achieve the convergence condition formulated in 8 (Main Loop). As seen from the figure, the clusters are initially assigned randomly. Next, the nested loop was computed 83 times to satisfy the condition in equation 4. During this iteration, pixels with low correlation are transferred between clusters to reduce the intra-cluster variation and achieve cluster homogeneity. Further, the nested loop is applied 56 times in order to satisfy the convergence condition mention in Equation 8. Intermediate outputs obtained after applying Equation 5, 6, 7 and 8 are presented for selected nested loop iterations. Finally, the final output obtained after convergence criteria are met as shown in Fig. 2.

III. EXPERIMENTAL RESULTS

The performance of the proposed data clustering method is presented on four datasets chosen from the field of image analysis, bioinformatics, remote sensing, and the stock market respectively. The details of experimentation along with quantitative as well as qualitative results including a comparison with seven state of the art algorithms namely K-Means (KM), Moving K-Means (MKM), Fuzzy C-Means (FCM), Adaptive Moving K- Means (AMKM), Enhanced Moving K-Means-1 (EMKM1), Enhanced Moving K-Means-2 (EMKM2) and Kernel K-Means clustering is presented. (Note: In all the experiments, RBF Kernel is used in K-Means clustering.)

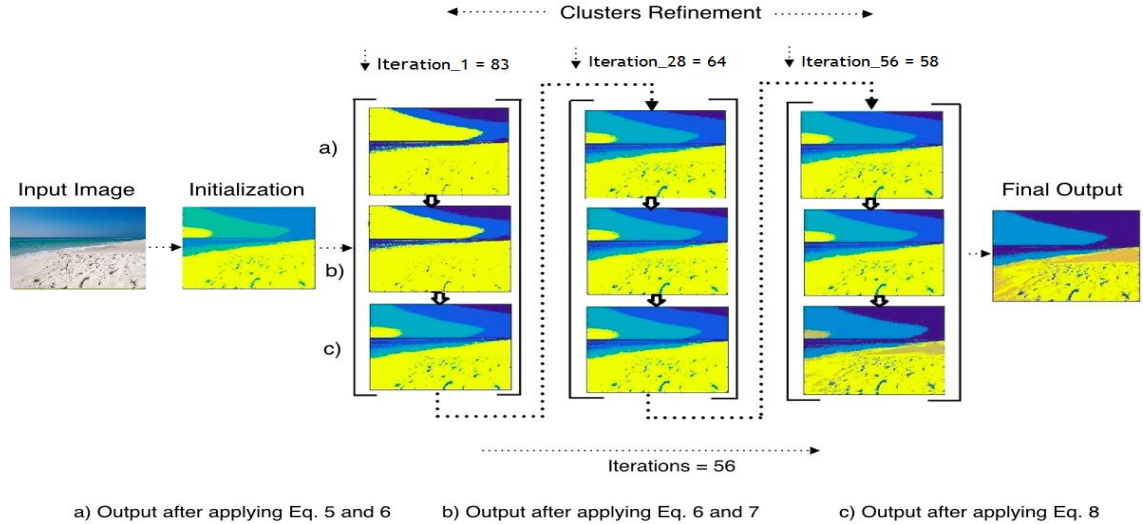

Fig. 2. Illustration of proposed VMKM algorithm with the number of iterations for main and nested loop

*A. Image Segmentation*

The performance of the proposed algorithm is computed on 100 images randomly selected from the Berkeley image dataset [4]. Each image in the selected 100 image dataset is segmented into five ($K = 5$) disjoint clusters. The segmentation results are demonstrated on four images of a boat, church, wall, and store as shown in the first row in Fig. 3. Quantitative and qualitative results along with a comparison with the state-of- the-art algorithms are presented for the sample images as well as the whole selected dataset.

*1) Qualitative Analysis:*

Qualitative Analysis is based on human visual sensitivity and is an indication of the algorithm's image segmentation efficiency. The first row in Fig. 3 shows the original images along with the segmentation results of the proposed VMKM algorithm and all other comparative clustering methods (K-Means, MKM, AMKM, FCM, Kernel K-Means, EMKM-1 and EMKM-2.

Figure 3(a) contains a boat with two ropes attached to its main sail and a bridge in the background. Conventional methods (K-Means, MKM, and FCM) and EMKM-1 show poor segmentation results as they fail to segment the bridge clearly and the ropes are not visible. AMKM and EMKM-2 have been partially successful but do not overall produce a clear segmentation. The ropes and the railings are best observed in Kernel K-Means and VMKM output.

Furthermore, Fig. 3(b) depicts a church with a dome, sky and few wires in the background. K-means, MKM, and EMKM-2 fail to segment the dome and the sky. Kernel K-Means was able to segment the dome but failed to segment the sky as well as the wires. EMKM-1 and FCM successfully segmented the dome and sky but could not segment the wires from the background. All three wires along with the regions of interest (dome, sky, and church) are clearly visible in VMKM output.

Next, Fig. 3(c) shows another example in which two persons are standing in front of a wall with a definite pattern and texture. K-Means shows inferior results as it fails to segment the faces of both persons and the texture on the wall. It was even unable to isolate the wine-glass from the shirt and fails to capture the patterns on the woman's dress. The pattern on the wall is not at all visible in MKM, and EMKM-1 while slightly observable in AMKM and FCM. Furthermore, the faces of both individuals including the pattern on the wall are best captured by only EMKM-2, Kernel K-Means and VMKM algorithms.

Finally, Fig. 3(d) consists of various vegetables with their prices displayed on the boards. K-Means, EMKM-1, Kernel K-Means, and EMKM-2 produce unfavorable results. They do not segment the text on the display boards and the cauliflowers are also not clearly seen. MKM, AMKM, VMKM, and FCM produce better results by segmenting the regions of interest, however only VMKM and FCM are able to differentiate the '9' in the right display board from the background. The texture of the vegetables is more prominent in VMKM. Overall, based on qualitative analysis, the proposed VMKM and Kernel K-Means algorithm produce optimum clusters as compared to other algorithms.

*2) Quantitative Analysis:*

Quantitative Analysis is a numeric method independent of human perception and errors which further assigns a definite numerical value to each algorithm. The clustering methods are applied to the images that are complex in terms of texture, vibrancy and have entities with irregular boundaries. Good segmentation is produced when data elements of a class within an image are assigned to the same cluster, thus producing minimum fitness for each class. Minimum Square Error (MSE) is a measure of the distance between the data elements and its centroid can be given by the following equation

$$MSE = \frac{1}{N}\sum_{j=1}^{K}\sum_{i \in c_j} ||v_i - c_j||^2 \quad . \quad (11)$$

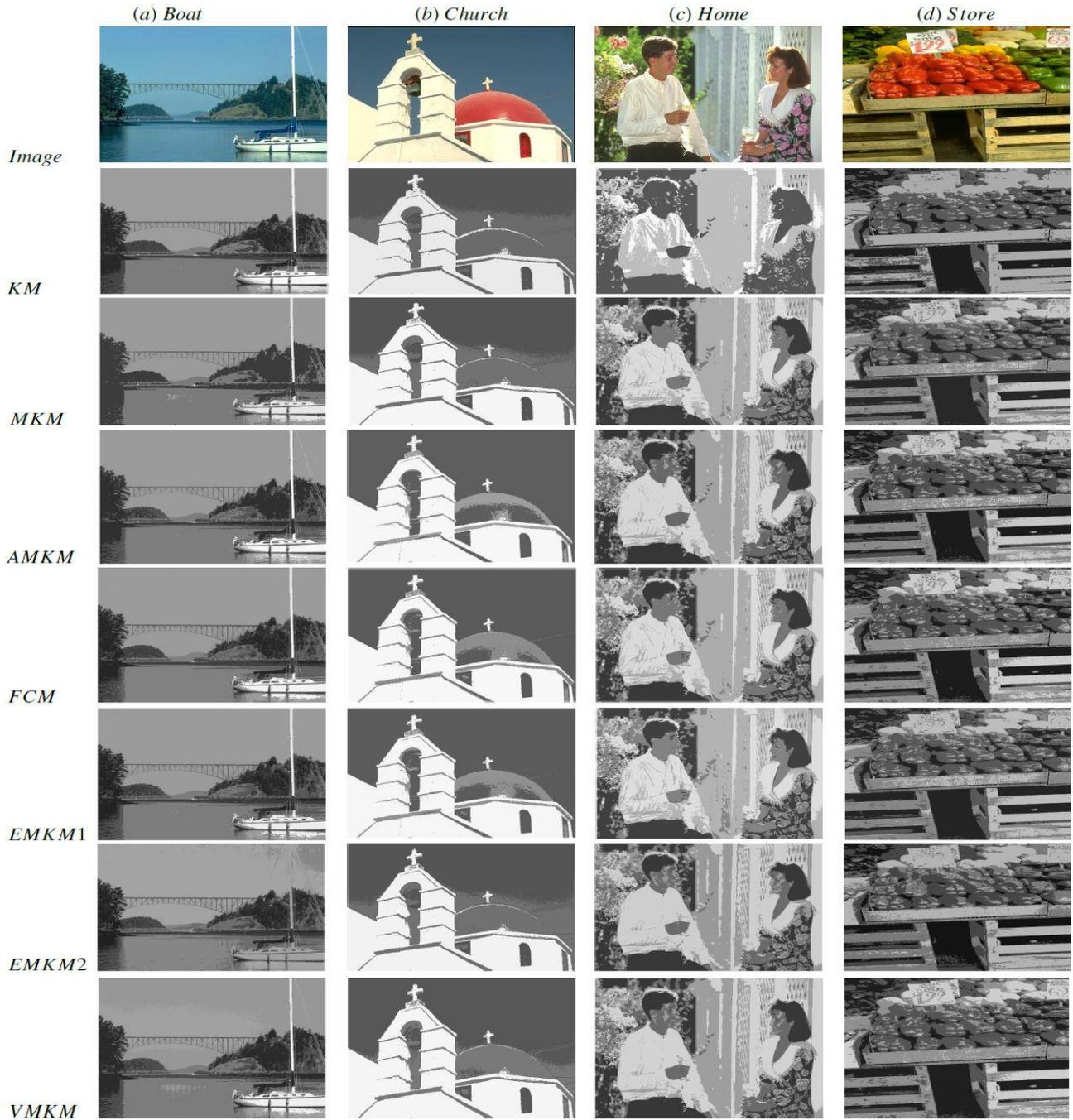

Fig.3. First row: original images. Segmentation results presented in second row: KM, third row: MKM, fourth row: AMKM, fifth row: FCM, sixth row: EMKM-1, seventh row: EMKM-2 and eighth row: VMKM.

In Eq. 11, *N* is the total number of pixels in an image, *K* is the desired number of clusters and $v_i$ is the grey scale intensity value of the $i$ pixel belonging to the $j^{th}$ cluster. The lower value of MSE represents assignment of data points to the most appropriate cluster.

Quantitative analysis of 100 images taken from the Berkeley Image Dataset is presented as Mean Squared Error Graph and Box Plot as shown in Fig. 4 and Fig. 5 respectively. The graph shows the MSE value trend in decreasing order for each image for all algorithms. It is clear from Fig. 4 that EMKM-2 produces the highest MSE while Kernel K-means and VMKM produces the lowest value for image dataset. This is also shown by the trend of the graph as VMKM and Kernel K-means are below all other algorithms.

Table I and the box plot as shown in Fig. 5 and summarize the performance of the algorithms based on their median and mean MSE values. A dotted line is used as a reference to compare the median MSE values. Kernel K-Means and

VMKM algorithm produce similar lowest MSE median value The mean MSE values for the images are tabulated in Table I and the lowest values are highlighted. We observe that the lowest MSE values are obtained again for Kernel K-Means and VMKM algorithm for the image dataset. This verifies the previously made inferences.

*3) Initialization Sensitivity Analysis*

The comparison graph is plotted to portray the initialization sensitivity of different algorithms. Twenty sets of randomly generated (i.e. using randperm command in Matlab) center values are generated for all the algorithms to segment the Boat image. The comparative evaluation in terms of mean MSE value is plotted against the number of tests as shown in Fig. 6. The comparison depicts that KM algorithm has the highest sensitivity to center value initialization. Further, Fig. 6 shows the higher sensitivity of MKM, and AMKM to initialization as compared to EMKM-1 and EMKM-2 algorithms. FCM, Kernel K-Means and VMKM have the least sensitivity to initialization as compared to other algorithms and they consistently produce good image segmentation performance as proven by the accompanying qualitative analysis. Kernel K-Means and VMKM produces the best image segmentation with similar lowest MSE.

*B. Bioinformatics*

In this experiment, we aim to group similar patterns using the proposed VMKM algorithm in the yeast bioinformatics dataset [7]. The dataset contains 6400 gene expression values of saccharomyces cerevisiae recorded during the metabolic change from fermentation to respiration observed using DNA microarrays. As the data set is large and consists of genes with small expression values, we trim the data set by keeping only those genes that show drastically changing expression profiles during the diauxic shift. The dataset is further refined by assigning one to the rows with yeast values having a variance greater than the 10th percentile while those below this value are represented by zero.

The VMKM algorithm is applied to cluster the data set into sixteen gene expression profiles. The quantitative results for gene expression classification measured on the basis of MSE are presented in Table I. It is evident from the table, that FCM and K-Means have the highest MSE values. All other algorithms including MKM, AMKM, EMKM-1 and EMKM-2 produce similar mid-range MSE values while the best clustering results are produced by VMKM and Kernel K-Means algorithm with similar lowest MSE with VMKM producing a slightly lower MSE value.

*C. Remote Sensing*

The remote sensing dataset used is a four-band multispectral image of Mysore district located in the southern part of India [13]. The image has dimensions of 1375 X 5929 pixels, area of 2.748km X 7.973 km and a resolution of 2.4m. The image records information about four crop classes (sugarcane, ragi, paddy, and mulberry) from different fields in the area. Multi-temporary imagery facilitates classification and identification of crops by taking into account changes in reflectance as a function of crop type. This information allows individuals to track positive and negative dynamics of crop development along with multiple natural processes such as erosion, soil properties, and vegetation conditions that can help government officials to plan for efficient land usage and crop management. In this study, we aim to classify the satellite image into four crop classes using the proposed VMKM clustering method. Quantitative results for crop classification are presented in Table I. The proposed method is also compared with seven state of the art clustering methods. It is evident from the table, that K-Means produces the worst clustering with the highest MSE while FCM, AMKM and MKM show relatively better results with smaller MSE values. All other algorithms (Kernel K-Means, EMKM-1, EMKM-2, and VMKM) produce relatively similar MSE values; however, the best clustering is produced by VMKM with the lowest MSE.

*D. Stock Market*

In this study, we aim to identify the stocks with higher chance of investment returns using the Istanbul stock dataset comprising of 30 stocks of different companies listed on Nasdaq[11]. Everyday absolute value and variance of the derivative of the stock price are used to define profitability. VMKM algorithm is applied to cluster all the stocks that have a larger variance in five clusters ($K = 5$) and hence a better chance of a positive return. Quantitative results for stock identification are presented in Table I. The proposed method is also compared with seven other popular clustering methods. From the table it is observed that K-Means produces the highest MSE followed by AMKM, MKM, EMKM-2, EMKM-1 and FCM respectively. The best clustering is produced by Kernel K-Means and VMKM with similar lowest MSE.

TABLE I. AVERAGE MSE FOR DIFFERENT ALGORITHMS APPLIED TO FOUR DATASETS

| Dataset | Average Mean Square Error | | | | | | | |
|---|---|---|---|---|---|---|---|---|
| | KM | MKM | AMKM | FCM | Kernel | EMKM-1 | EMKM-2 | VMKM |
| Image Analysis | 175.2 | 156.2 | 170.7 | 191.70 | 139.4 | 158.2 | 196.1 | 142.5 |
| Bioinformatics | 99.1 | 8.5 | 12.4 | 112.8 | 8.4 | 9.8 | 9.9 | 8.2 |
| Remote Sensing | 95.2 | 64.1 | 61.3 | 67.1 | 44.8 | 47.5 | 48.9 | 44.0 |
| Stock Market | 197.7 | 150.3 | 182.4 | 116.6 | 89.5 | 139.4 | 141.3 | 89.4 |

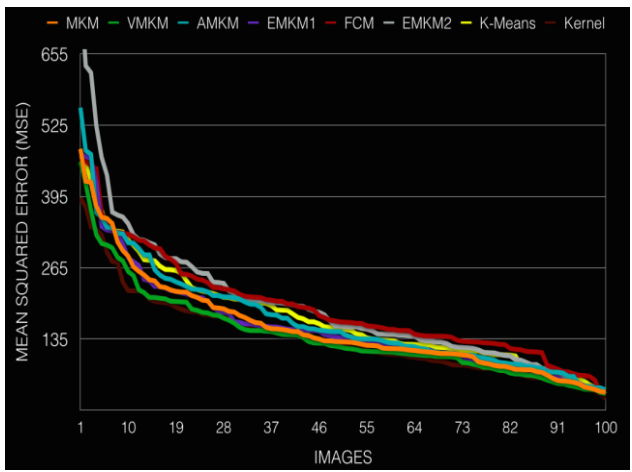

Fig.4. Graph comparing average MSE of different algorithms applied on the image dataset.

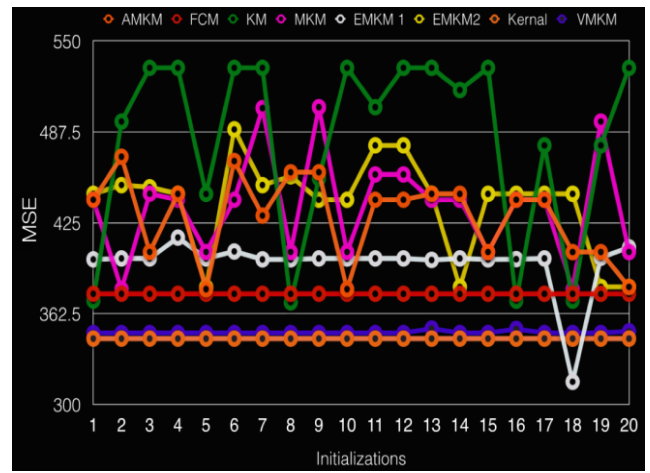

Fig. 5. Box plot for segmentation variances with median MSE (indicated with pink horizontal line) for all algorithms applied on the image dataset.

*E. Comparison of VMKM with Kernel K Means*

It is evident from Table I that Kernel K-Means performs almost as well VMKM for bioinformatics, remote sensing, and stock market and outperforms the proposed algorithm for image analysis. Despite the comparable performance of Kernel K-Means to VMKM and ability to separate non-linear data, it has some obvious disadvantages. Former is a complex algorithm with large time complexity. Also, knowledge about the kernel to be used for the data separation is essential before applying Kernel K-Means. VMKM doesn't have any such limitations making it a superior choice as compared to Kernel K-Means.

IV. CONCLUSION

In this paper, a new version of the MKM algorithm known as Variance based Moving K-Means (VMKM) is presented. The process of assigning and then transferring data elements takes into account the cluster variance and correlation of data elements with the parent as well as neighbouring clusters. This results into lower intra-cluster variance with no dead centers. The algorithm also has no sensitivity to initialization further making it an optimal clustering choice. The algorithm was effectively able to cluster useful information from four datasets as demonstrated with the lowest mean and median MSE.